\documentclass[10pt,twocolumn]{article}

\usepackage[utf8]{inputenc}      % UTF-8 input
\usepackage[T1]{fontenc}         % T1 font encoding
\usepackage{lmodern}             % Latin Modern fonts
\usepackage{amsmath, amssymb}    % math symbols
\usepackage{graphicx}            % figures
\usepackage{hyperref}            % hyperlinks
\usepackage{natbib}              % bibliography
\usepackage{geometry}            % adjust margins
\usepackage{booktabs}
\usepackage{multirow}
\usepackage{algorithm}
\usepackage{algorithmic}
\title{\textit{SynthGenNet}: a self-supervised approach for test-time generalization using synthetic multi-source domain mixing of street view images}

\author{
Pushpendra Dhakar$^{a}$ \quad Prachi Chachodhia$^{a}$ \quad Vaibhav Kumar$^{a*}$ \\
$^{a}$GeoAI4Cities Research Lab, Department of Data Science and Engineering, IISER Bhopal\\
\texttt{dhakarpushpendra44@gmail.com, prachi23@iiserb.ac.in, *vaibhav@iiserb.ac.in}
}

\date{} 

\begin{document}

\maketitle

\begin{abstract}
Unstructured urban environments present unique challenges for scene understanding and generalization due to their complex and diverse layouts. We introduce \textit{SynthGenNet}, a self-supervised student-teacher architecture designed to enable robust test-time domain generalization using synthetic multi-source imagery. Our contributions include the novel ClassMix++ algorithm, which blends labeled data from various synthetic sources while maintaining semantic integrity, enhancing model adaptability. We further employ Grounded Mask Consistency Loss (GMC), which leverages source ground truth to improve cross-domain prediction consistency and feature alignment. Pseudo-Label Guided Contrastive Learning (PLGCL) mechanism is integrated into the student network to facilitate domain-invariant feature learning through iterative knowledge distillation from the teacher network. This self-supervised strategy improves prediction accuracy addressing real-world variability, bridging the sim-to-real domain gap, and reliance on labeled target data, even in complex urban areas. Outcomes show our model outperforms the state-of-the-art (relying on single source) by acheiveing ~50\%  Mean Intersection-Over-Union (mIoU) value on real-world datasets like Indian Driving Dataset (IDD). 
\end{abstract}

\section{Introduction}
Semantically Segmented (SS) maps of Street View Images (SVIs) play a key role in various computer vision applications, including fields like autonomous driving \cite{Liu2024} and urban planning \cite{Jia2024}. These applications require precise pixel-level classification, as inaccuracies can impact the overall effectiveness of the system. Despite substantial advancements, creating robust SS maps in diverse and unstructured environments remains challenging due to real-world dynamism. The challenge is especially pronounced in the urban landscapes of developing nations, such as India. While ample labeled data could enhance the accuracy of current SS approaches, obtaining environment-specific labeled training data is often impractical. 
Addressing the challenge of training models on a source domain that can be generalized effectively to unseen target domains, i.e., Domain Generalization (DG), is a potent solution to the discussed challenge. DG reduces the dependency of SS tasks on labeled target data \cite{bi2024learning, peng2021global} making it suitable for applications that are implemented in extremely diverse and unseen environments, which is also the focus of our paper. 

Apart from effective feature learning, the efficacy of the model to extract domain invariant features is also dependent on the quality of source data, which is scarce. By training models on pixel-perfect synthetic SS maps generated for desired environments, invariant features can be learned. However, due to sim (source)-to-real (target) domain gaps, existing due to art effects and noise is another challenge that makes accurate generalization tasks difficult.  
Apart from the discussed challenge the current body of work utilizes a single-source domain which is often insufficient for adapting to complex, dynamic urban settings, where factors like unstructured scenes, extreme weather and varying lighting are present. To address these limitations our work proposes utilizing multi-source synthetic datasets and apply an efficient mixing strategy in the learning process. 

Recently, some Unsupervised Domain Adaptions (UDA) techniques such as multi-source-domain data-mixing strategies like CutMix \cite{yun2019cutmix} and ClassMix \cite{olsson2021classmix}, Masked Image Consistency(MIC)\cite{hoyer2023mic} and Contrastive Learning Framework \cite{basak2023pseudo} have been proposed and, are employed through self-supervision learning for test-time generalization in our hypothesis. These approaches combine features from multiple domains to generate synthetic samples, encouraging models to learn domain-invariant representations. The algorithms preserve semantic class integrity by overlaying class-specific segments from one image onto another, maintaining realistic boundaries and spatial relationships. This promotes context-aware feature learning, which is especially useful in structured environments like cityscapes, where objects such as roads, buildings, and vehicles have predictable placements. 

However, these approaches are less effective in highly unstructured or chaotic scenes, as they rely on relatively structured source domains to create coherent mixed samples. To overcome this limitation, we propose a new mixing strategy, ClassMix++ which learns the mixing of samples from multiple synthetic source images. We hypothesize that mixing from multiple source images can enhance the ability of the model to generalize by using multiple diverse features and variations, making it more robust to unseen environments.
ś
To enhance the generalizability of mixing algorithms through learning invariant features, we introduce Grounded Mask Consistency (GMC) loss. The algorithm is inspired by MIC \cite{hoyer2023mic} which guides a model to handle missing or occluded information by predicting masked portions of the target image. GMC overcomes the limitation of MIC, i.e., being sensitive to noise from pseudo-labels, especially in domains with high variability, as these labels can introduce errors when applied to masked regions. 
Instead of pseudo label-based learning, we use GMC on labeled source images to enforce spatial coherence by training the model to make reliable predictions even when portions of the input are unclear, which is useful for recognizing small or overlapping objects in complex environments. 
The loss is implemented in the teacher network, which trains on multi-source labeled data to learn domain-invariant features.

To achieve generalization on unseen target domains at test time, we propose a self-supervised learning approach within our architecture, \textit{SynthGenNet}. Our method draws inspiration from Unsupervised Domain Adaptation (UDA) approaches \cite{hoyer2023mic, olsson2021classmix, marsden2022contrastive} and is designed for effective test-time generalization, also refered as Generalization by Adaptation (GA) \cite{ambekar2023learning}.  
\textit{SynthGenNet} overcomes limitations of traditional DG methods by improving data diversity, enhancing label consistency, and aligning feature representations across multiple domains. Our architecture leverages a trained teacher network to generate pseudo-labels for target data, guiding the student network in refining its segmentation capabilities on previously unseen domains. The student network iteratively incorporates these pseudo-labels to enhance feature discrimination, enabling more precise feature alignment across diverse domains. We demonstrate the efficacy of approach on street-view images captured in cities from developing nations, which pose unique challenges due to their unstructured and highly diverse urban landscapes compared to planned urban environments. The primary contributions and highlights are as follows: 

 \begin{itemize}

\item A novel self-supervised learning architecture, \textit{SynthGenNet}, that enables test-time generalization on unseen domains by leveraging synthetic multi-source images. 
\item A novel ClassMix++ algorithm, which employs labeled data from multiple synthetic sources to create a diverse mixed dataset, preserving semantic boundaries and contextual coherence across varied domains. 
\item A novel GMC loss, utilizing source ground truth labels to enhance model prediction consistency and feature learning across domains. 
\item Implementation of Pseudo-Label Guided Contrastive Learning (PLGCL) \cite{basak2023pseudo} in the student network, encouraging feature alignment across domains and promoting the learning of domain-invariant features. During training, the student network distills knowledge from the teacher network, iteratively refining pseudo-labels for precise predictions on unseen target data. This approach enhances the architecture's robustness in variable real-world conditions and addresses the sim-to-real domain adaptation gap. 
 
\end{itemize}

\section{Related Work}
\begin{figure*}[!t]
    \centering
    \includegraphics[width=12cm, height=4.5cm]{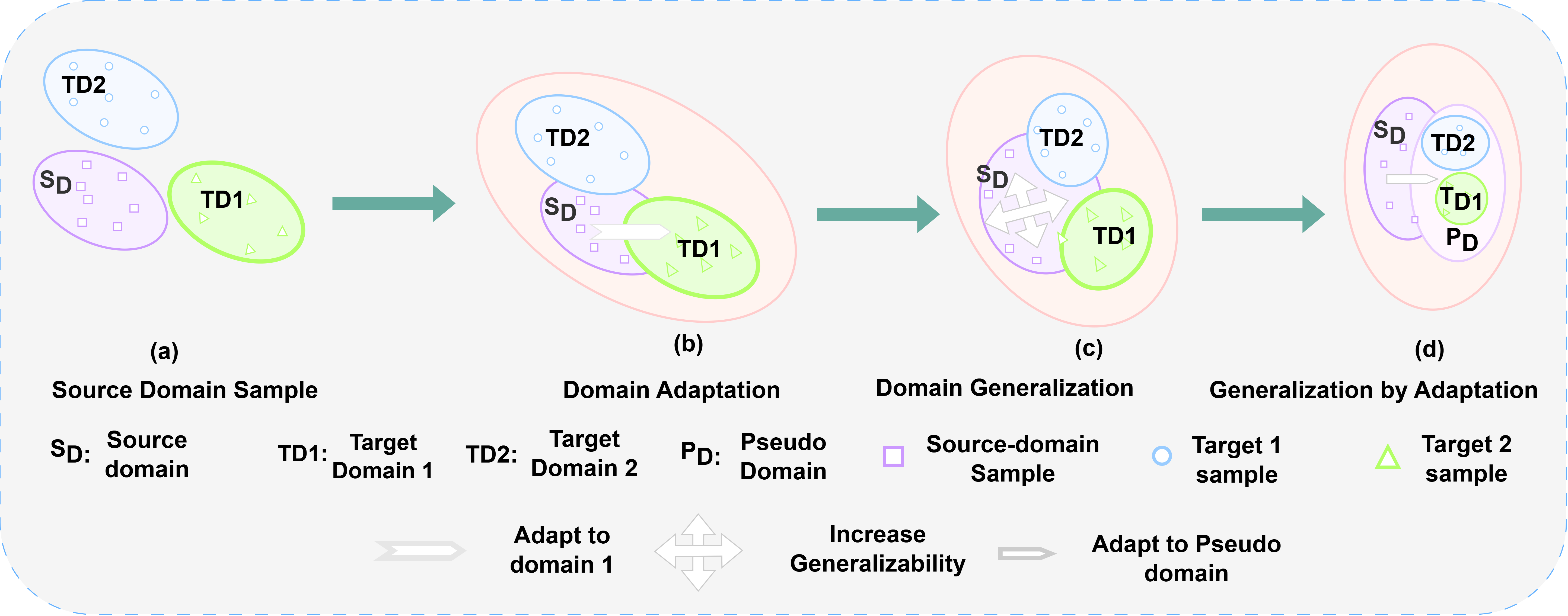}
    \caption{Representation of transition from Domain Adaptation (DA) to Generalization by Adaptation (GA) 
(a) shows the initial distributions of the Source Domain (SD) and Target Domains (TD1 and TD2). (b) Domain adaptation applied to TD1 helps in reducing the gap between the domains. (c) Domain Generalization (DG) improves model robustness for previously unseen domains. (d) Adaptation-based generalization creates a pseudo-domain (PD) which encompasses TD1 and TD2 and helps in better generalization.}
    \label{figure 1}
\end{figure*}
\textbf{Unsupervised Domain Adaptation (UDA)} is a crucial strategy for bridging the performance gap between labeled source and unlabeled target domains. However, the distributional discrepancy between source and target domains can lead to significant performance drops \cite{hoyer2022daformer,hoyer2023mic,hoyer2022hrda}. Early works focused on aligning feature distributions between domains\cite{bai2024prompt,jin2020feature}
often use adversarial learning \cite{li2018domain,park2024adversarial}
to make feature distributions indistinguishable. This method encourages the extraction of domain-invariant features, improving generalization on the target domain.
Recent advances have introduced self-supervised learning \cite{sun2019unsupervised}
techniques that generate pseudo-labels for the target domain, but they often rely on fully annotated source datasets, raising privacy concerns. Source-free UDA \cite{klingner2022unsupervised} has come up to solve this problem by letting models adapt without access to source data and making UDA more useful by making sure that models can work well in new target domains while maintaining privacy.

\textbf{Domain Generalization (DG)} aims to generalize models trained on source domains to unseen target domains. It is more challenging than UDA, which uses target domain data during training. Early DG approaches focused on learning domain-invariant features that remain consistent across varying source domains. Methods like Maximum Mean Discrepancy (MMD) \cite{chen2024domain,ding2022domain} 
and contrastive losses \cite{basak2023pseudo,jeon2021feature} 
were introduced to minimize domain gaps. Recent work has explored meta-learning \cite{khoee2024domain}
 and ensemble learning \cite{mi2023multiple}
 as strategies to address domain shifts. But to achieve robust domain generalization remains still a complex challenge, and while methods like AugFormer \cite{schwonberg2023augmentation} 
and SHADE  \cite{zhao2022style} have shown promising results, but more research is still required to overcome the multifaceted challenges of domain DG for unseen complex domains.

\textbf{MIC Loss} \cite{hoyer2023mic} builds upon advances in masked image modeling from self-supervised learning but focuses on enhancing domain adaptation through prediction rather than representation learning. It uses a teacher-student model updated with the Exponential Moving Average (EMA) to generate pseudo-labels for the target domain \cite{hoyer2023mic}. The teacher model provides high-quality pseudo-labels, while the student learns from masked input to extract relevant features. To handle errors, MIC applies a quality weight based on softmax probabilities.

\textbf{Contrastive Learning (CL)} is a technique that improves representation learning tasks like image classification \cite{jiang2024cbda} 
by pulling positive sample pairs closer and pushing negative pairs apart. This technique has been applied to improve pixel-level features in SS. Recent approaches have also explored prototype and distribution-wise methods for segmentation \cite{ahn2024style}\cite{basak2023pseudo}. In this study, we used contrastive loss on the student network with pseudo-labels to deal with problems in domain generalization during simulated to real generalization.

\section{Methodology}
\label{sec:SynthGenNet}

\textit{SynthGenNet} is a teacher-student based architecture that helps to address the limitations of DG methods presented in (see \textbf{Figure} \ref{figure 1}), by using multi-source synthetic data to address domain shifts, particularly in complex urban scenes. The teacher network is initially trained on multi-source domain data, including synthetic samples generated using the ClassMix++ algorithm. The use of classmix++ helps to create diverse mixed datasets. Whereas, GMC loss reinforces contextual relationships to ensure consistent segmentation across diverse domains. 

\begin{figure*}[!t]
    \centering
    \includegraphics[height=7cm, width=15cm]{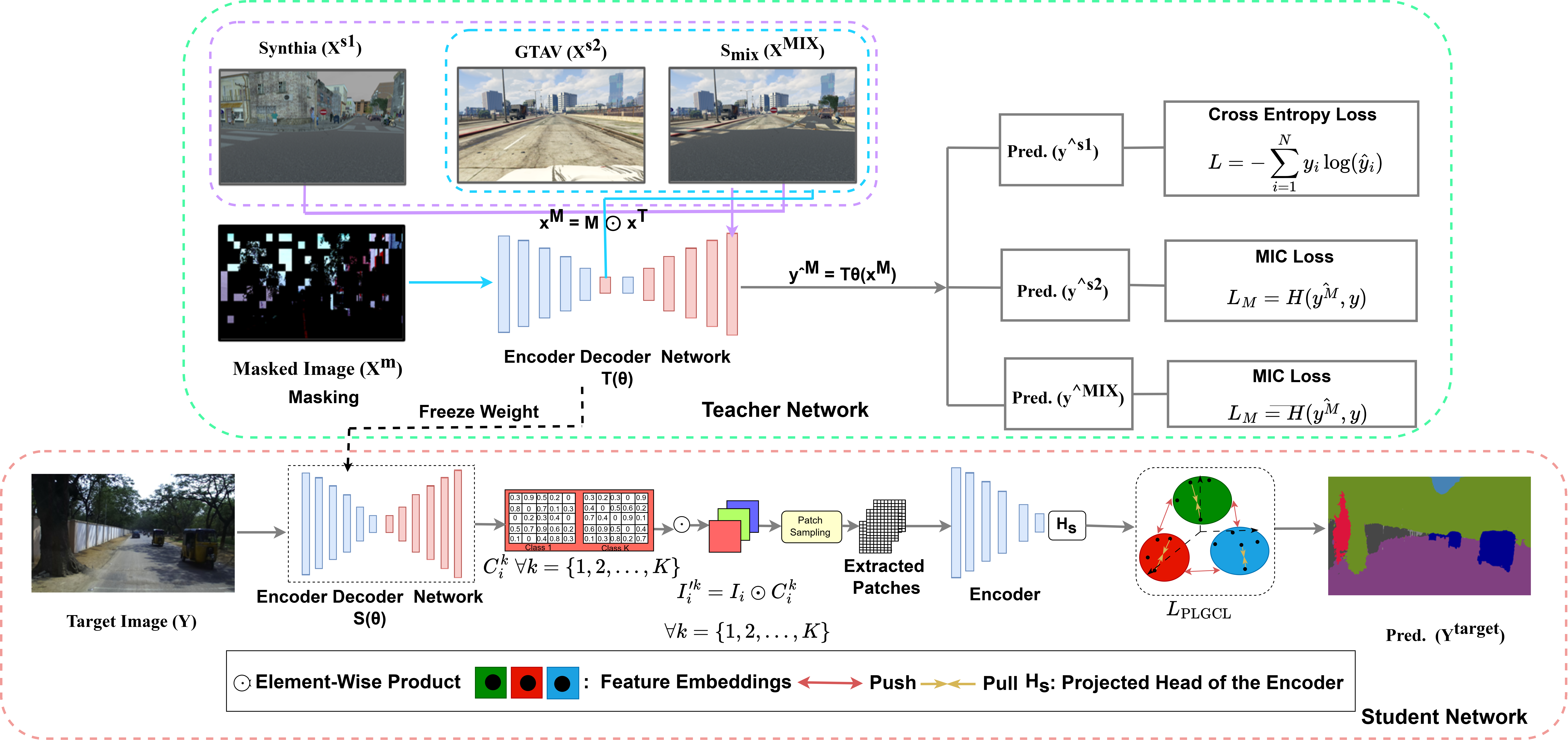}
    \vspace{0.1cm}  
    \caption{Training workflow for SynthGenNet. The teacher network is trained with original and augmented source data using GMC loss for robust feature learning. Pseudo-labels generated by the teacher guide the student network on target domain images.}
    \vspace{0.1cm} 
    \label{figure 2}
\end{figure*}

The encoder-decoder structure of DeepLabV3+ is used in both the teacher and student networks due to its efficiency in capturing both local and global features, which aids in faster processing. ResNet \cite{he2016deep} is used as a backbone to extract feature maps from the input image, while the decoder refines these features to produce detailed segmentation maps. The architecture of \textit{SynthGenNet}, depicted in \textbf{Figure}~\ref{figure 2}. Student network then applies PLGCL loss, aligning similar features across domains while minimizing domain-specific biases. This architecture then helps \textit{SynthGenNet} to maintain reliable segmentation performance across diverse and challenging environments.

\subsection{Proposed Classmix++ Algorithm} \label{subsec:classmix++}
\begin{figure}
    \centering
    \includegraphics[width=1\linewidth]{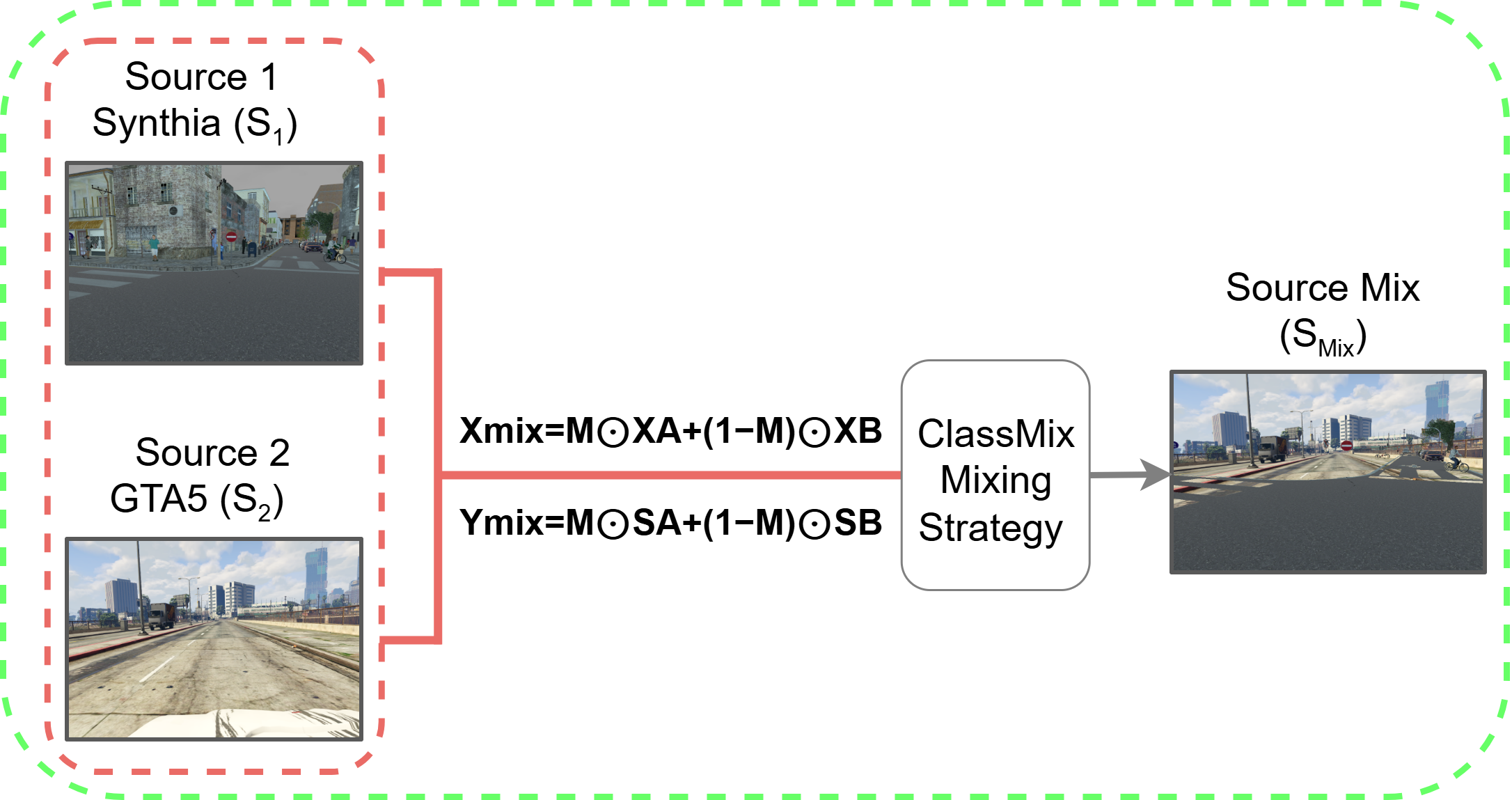}
    \caption{Classmix++ implementation strategy}
    \label{fig-3}
\end{figure}

Classmix++ uses ground truth to guide the generation of a binary mask to overcome the limitations of classmix.  The detailed steps for implementing ClassMix++ are provided in \textbf{Algorithm}~\ref{alg:classmix}, In \textbf{Step 1}, two labeled images, \(A \) and \(B \), are selected, each paired with corresponding ground truth annotations \( S_A \) and \( S_B \). In \textbf{Step 2}, an \textit{argmax} operation is applied to the ground truth masks, producing pixel-wise class predictions by identifying the dominant class at each pixel location. \textbf{Step 3} involves selecting a subset of semantic classes from those present in \( \tilde{S}_A \) to form a binary mask, determining which regions of \( A \) and \( B \) will be mixed. Specifically, pixels from \( A \) belonging to the selected classes are combined with the remaining pixels from \( B \). \textbf{Figure}~\ref{fig-3} illustrated the strategy of Classmix++
In \textbf{Step 4}, a binary mask \( M \) is created based on the selected classes. For each pixel location M(i,j) is set to 1 if the class is at that location in \(\tilde{S}_A \) belongs to the selected subset; otherwise, it is set to 0.
At last, in \textbf{Step 5}, the mixed image \( X_{\text{mix}} \) is produced by combining pixels from \( A \) and \( B \) and in \textbf{Step 6}, the algorithm returns the newly generated mixed samples.

These steps ensure that generated images and labels remain realistic and meaningful after augmentation. Classmix++ allows the model to learn from a larger and more diverse dataset, which helps \textit{SythGenNet} to improve its performance. This advanced strategy focuses on semantic consistency to understand object relationships and properties, which makes it effective in generalization scenarios. 

\begin{algorithm}[t]
    \caption{\textbf{ClassMix++ Algorithm}}
    \label{alg:classmix}
    \begin{algorithmic}[1]
        \REQUIRE Two labeled samples $A$ and $B$
        \ENSURE Mixed samples $X_A$ and $Y_A$
        \STATE $S_A \leftarrow (A)$
        \STATE $S_B \leftarrow (B)$
        \STATE $\tilde{S}_A \leftarrow \arg\max_{c'} S_A(i, j, c_0)$ \COMMENT{Pixel-wise argmax over classes}
        \STATE $C \leftarrow$ Set of different classes in $\tilde{S}_A$
        \STATE $c \leftarrow$ Randomly select subset of $C$ such that $|c| = |C|/2$
        \FORALL{$i, j$}
            \STATE $M(i, j) \leftarrow \begin{cases} 
                1, & \text{if } \tilde{S}_A(i, j) \in c \\
                0, & \text{otherwise}
            \end{cases}$
        \ENDFOR
        \STATE $X_A \leftarrow M \odot A + (1 - M) \odot B$ \COMMENT{Mix images}
        \STATE $Y_A \leftarrow M \odot S_A + (1 - M) \odot S_B$ \COMMENT{Mix predictions}
        \STATE \textbf{return} $X_A, Y_A$
    \end{algorithmic}
\end{algorithm}

\subsection{Poposed GMC Loss for Feature Stability} \label{subsec:GMC Loss}
MIC\cite{hoyer2023mic} module is a new approach in UDA that enhances visual recognition by learning spatial context relations of the target domain. MIC enforces consistency between predictions of masked target images and pseudo-labels generated by an exponential moving average teacher. While the MIC module effectively learns spatial context by enforcing prediction consistency, its reliance on pseudo-labels can limit its accuracy, particularly in complex or visually ambiguous target domains. To address this, we introduce GMC loss, which integrates the ground truth of synthetic labels instead of pseudo-labels, enhancing contextual learning and reducing error for improved target domain adaptation. By focusing on both content and context, GMC loss allows the model to generalize more effectively between training data and real-world applications. In the  process of generating the \textit{masked image} a patch mask \( M \) is randomly sampled from a uniform distribution \( U(0, 1) \), as defined by \textbf{Equation}~\eqref{eq:mask_def}:

\begin{equation}
M_{mb+1:(m+1)b, nb+1:(n+1)b} = [v > r] \quad \text{with} \quad v \sim U(0, 1)
\label{eq:mask_def}
\end{equation}

Here, \( b \) is  \textit{patch size}, \( r \) is the \textit{mask ratio} and \( m \in [0..W/b-1] \), \( n \in [0..H/b-1] \) are the patch indices. The resulting \textit{masked image} \( x^M \) is obtained by applying element-wise multiplication of the mask with the original image.
The masked prediction \( \hat{y}^M \) is computed by passing the masked image \( x^M \) through the model, as shown in \textbf{Equation}~\eqref{eq:masked_pred}:

\begin{equation}
\hat{y}^M = f_\theta(x^M)
\label{eq:masked_pred}
\end{equation}

This setup challenges the model to rely on visible regions for segmentation, making it more robust to incomplete data in the target domain. To guide the model for utilizing contextual clues for accurate label reconstruction without having full access to the image, we used GMC loss \( L_M \). This loss function is defined as the cross-entropy between the \textit{masked prediction} \( \hat{y}^M \) and the \textit{ground truth labels} \( y \), as expressed in \textbf{Equation}~\eqref{eq:mic_loss}:

\begin{figure}[t]
    \centering
    \includegraphics[width=.8\linewidth, height=0.5\linewidth]{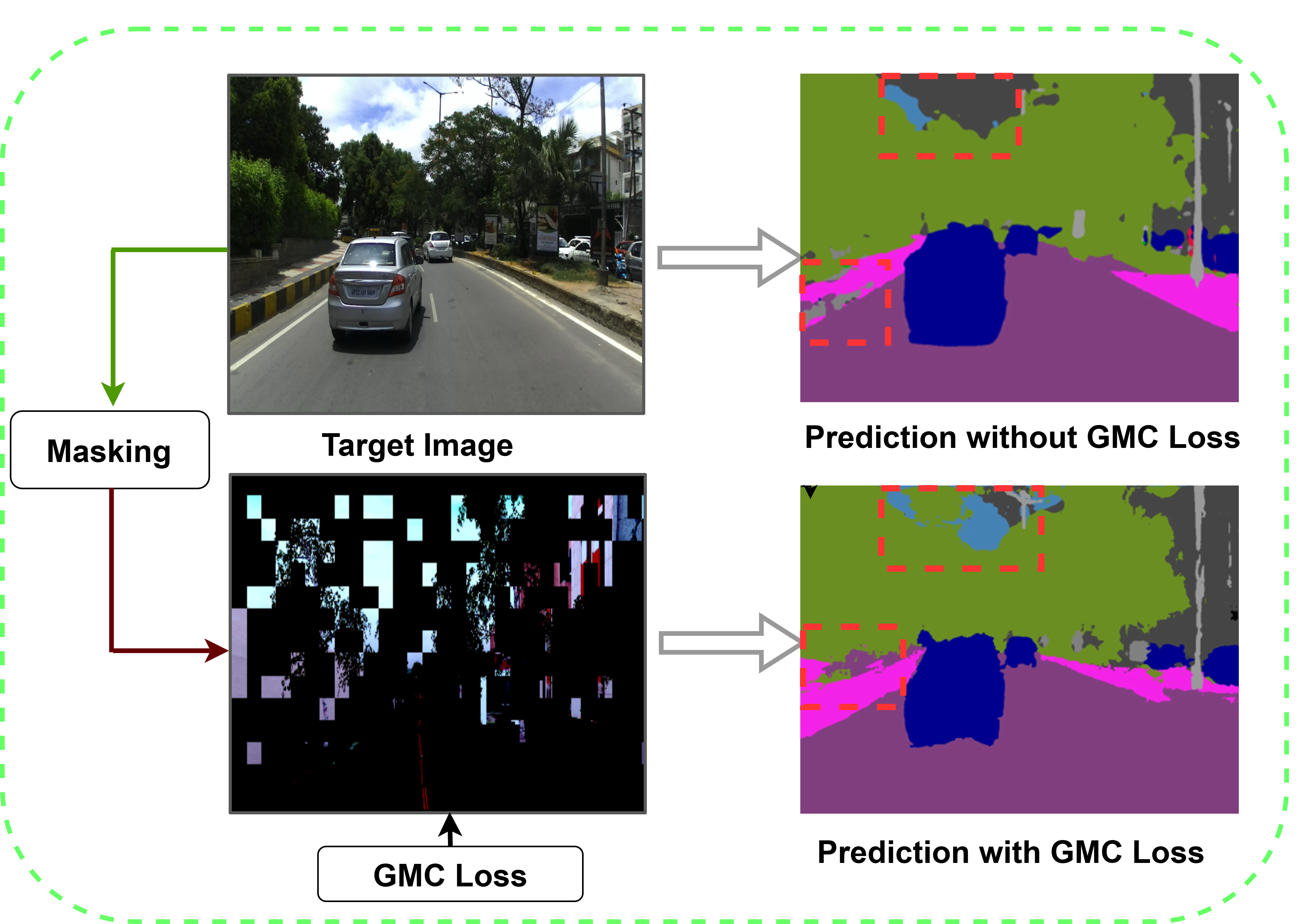}
    \caption{\textbf{Impact of GMC loss on segmentation} Left section shows the masking process applied to the target image, producing a partially obscured image used for training. The right side of the figure compares the predictions without GMC loss (top) and with GMC loss (bottom).  
     }

    \label{fig-4}
\end{figure}

\begin{equation}
L_M = H(\hat{y}^M, y)
\label{eq:mic_loss}
\end{equation}
Here, \( H \) represents the cross-entropy loss and \( y \) denotes the ground truth labels.
\textbf{Figure}~\ref{fig-4} shows the positive effect of applying \textit{GMC loss} on segmentation performance. Predictions with this loss are significantly more accurate, especially in challenging areas with occlusions or similar local appearances.

\subsection{Integration of PLGCL Loss for SythGenNet} \label{subsec: PLGCL Loss}
PLGCL \cite{basak2023pseudo} is integrated into the student domain, which can mine class-discriminative features without explicit training. It is a hierarchical approach, where multiple patches have been generated for each class present in a single image and a confidence score is assigned to each patch. The patch with a higher confidence score will represent the particular class. In our application 13 classes such as sidewalk, road etc. have been considered. This approach avoids class collision, which helps to handle unstructured complex environments like India.  

To get these patches for PLGCL, class-aware patch sampling has been used which involves creating pseudo-labels, which represent class confidence metrics for each pixel, identifying regions that are likely to belong to specific classes.
These pseudo-labels are then multiplied with the original image to generate attended images, from which patches are extracted. The sampling is guided by the average confidence of pixels within a patch and the average patch entropy, prioritizing patches containing objects or parts of a particular class.

In a mini-batch, let the \(i\)-th image be denoted as \( I_i \), which consists of \(M\) pixels, each identified as \( I_i(m) \) for \( m \in [1, M] \). A teacher-student network, parameterized by \( t\theta, s\theta \), was used to derive pseudo-labels \( Y'_i \) from the teacher network. These pseudo-labels were represented as class-confidence metrics \( C_i \), where \( C_i^k(m) \) denotes the likelihood of the \(m\)-th pixel in image \( I_i \) belonging to class \( k \). The confidence map is then element-wise multiplied with \( I_i \) to produce the attended image \( I'^k_i \), as shown in \textbf{Equation}~\eqref{eq:attended_image}:
\begin{equation}
I'^k_i = I_i \odot C_i^k
\label{eq:attended_image}
\end{equation}

Here, \( \odot \) represents the element-wise multiplication. The patches generated from this attended image are denoted as \( P^k_{i,j} \), representing the \( j \)-th patch of the \( i \)-th attended image for class \( k \). This whole process is called the patch sampling process. Proper patch selection is crucial for CL. Each anchor patch belonging to class \( k \), patches containing objects or parts from class \( k \) were labeled as positive, while patches from other \( K-1 \) classes were labeled as negative. The sampling of patches is based on class confidence, with the average confidence of a patch \( P^k_{i,j} \) defined in \textbf{Equation}~\eqref{eq:avg_confidence}:
\begin{equation}
\text{Avg}^k_{i,j} = \frac{1}{|P^k_{i,j}|} \sum_{m \in P^k_{i,j}} C^k_i(m)
\label{eq:avg_confidence}
\end{equation}

High average patch confidence \( \text{Avg}^k_{i,j} \) suggests that the patch is likely to contain objects or parts from class \( k \), whereas values closer to zero suggest the opposite. However, this confidence metric alone does not account for the intensity appearance or the uncertainty between classes. Therefore, the average patch entropy was also computed based on the attended image \( I'^k_i \). For a patch \( P^k_{i,j} \), the average entropy \( \text{Ent}^k_{i,j} \) is given in \textbf{Equation}~\eqref{eq:avg_entropy}:
\begin{equation}
\text{Ent}^k_{i,j} = \frac{1}{|P^k_{i,j}|} \sum_{m \in P^k_{i,j}} F(I'^k_i(m))
\label{eq:avg_entropy}
\end{equation}

Here,  \( F(x) \) is the entropy function, defined in \textbf{Equation}~\eqref{eq:entropy_function} and \( I'^k_i(m) \) represents the intensity value of the \( m \)-th pixel inside the patch \( P^k_{i,j} \). The entropy \( \text{Ent}^k_{i,j} \) captures three types of information for each patch \( j \) in image \( i \):

\begin{itemize}
\item[\textbf{.}] confidence of the patch belonging to class \( k \),
\item[\textbf{.}] uncertainty regarding the other \( K-1 \) classes,
\item[\textbf{.}] intensity appearance from the image \( I_i \).
\end{itemize}

\begin{equation}
F(x) = -x \log(x) - (1 - x) \log(1 - x)
\label{eq:entropy_function}
\end{equation}

By applying contrastive learning with patch sampling, the patches that have n-nearest \( \text{Ent}^k_{i,j} \) values to an anchor patch of class \( k \) values are marked as positive, while the other patches are marked as negative. These patches are then encoded by an encoder \( E_S \) and projected using a head \( H_S \) to obtain feature embeddings. The embeddings were then used for contrastive loss formulation. The embedding of the anchor point serves as the query and is contrasted with embeddings from other patches (considered as keys), which forms the foundation of the contrastive learning approach.
PLGCL, pseudo-labels \( Y'_U \) for the unlabeled set \( I_U \) and integrates class-wise patches for a more comprehensive evaluation. The \( u \)-th query patch of class \( k \) is denoted as \( P_u \), with the corresponding positive key patch \( P^+_v \) and negative key patch \( P^-_v \). Their embeddings \( f_u \), \( f^+_v \) and \( f^-_v \) are computed using \textbf{Equation}~\eqref{eq:embedding_computation}:
\begin{equation}
\{f_u, f^+_v, f^-_v \} \leftarrow H_S(E_S(\{P_u, P^+_v, P^-_v\}))
\label{eq:embedding_computation}
\end{equation}

These patches were encoded and projected to embeddings, enabling contrastive learning with InfoNCE loss \cite{oord2018representation}, to ensure similarity between similar target instances and their positive pairs while minimizing the similarity with the negative samples for better generalization (\textbf{Equation}~\eqref{eq:infonce_loss})

\begin{equation}
L = -\mathbb{E}_J \log \frac{\exp(f_u^T \cdot f^+_v / \tau)}{\exp(f_u^T \cdot f^+_v / \tau) + \sum_{k^-} \sum_v \exp(f_u^T \cdot f^-_v / \tau)}
\label{eq:infonce_loss}
\end{equation}

The closed-form upper bound for the loss is modeled as Equation~\eqref{eq:upper_bound_loss}. Here, the loss simplifies the calculations by approximating the expectation. 
\begin{equation}
\begin{split}
L \leq \log \mathbb{E}_J \left[ \exp\left(\frac{f_u^T \cdot f^+_v}{\tau}\right) \right.
\left. + \sum_{k^-} \sum_v \exp\left(\frac{f_u^T \cdot f^-_v}{\tau}\right) \right] \\
- \frac{f_u^T \cdot \mathbb{E}[f^+_v | k^+]}{\tau}
\end{split}
\label{eq:upper_bound_loss}
\end{equation}

Assuming that the embeddings \( f^+_v \) and \( f^-_v \) follow Gaussian distributions:
\begin{align*}
f^+_v &\sim \mathcal{N}(\mu^+_v, \sigma^+_v) \\
f^-_v &\sim \mathcal{N}(\mu^-_v, \sigma^-_v)
\end{align*}

The loss function has been further refined. These Gaussian assumptions help model the distribution of embeddings, where \( \mu^+_v \) and \( \sigma^+_v \) represent the mean and covariance of the positive samples, while \( \mu^-_v \) and \( \sigma^-_v \) represent those of the negative samples. The patch-wise pseudo-label guided contrastive loss \( L_{P_u}^{\text{PLGCL}} \) is expressed in Equation~\eqref{eq:plgcl_loss}:
\begin{equation}
\begin{split}
L_{P_u}^{\text{PLGCL}} = \log \Bigg[ &\exp \left( \frac{f_u^T \cdot \mu^+_v}{\tau} + 
\frac{\lambda}{2\tau^2} f_u^T \cdot \sigma^+_v f_u \right) \\
& + \zeta \sum_{k^-} \exp \left( \frac{f_u^T \cdot
 \mu^-_v}{\tau} + \frac{\lambda}{2\tau^2} f_u^T \cdot \sigma^-_v f_u \right) \Bigg] \\
& - \frac{f_u^T \cdot \mu^+_v}{\tau}
\end{split}
\label{eq:plgcl_loss}
\end{equation}

Here, \( \zeta \) scales the sum of negative embeddings. As training progresses, the statistics become more informative and \( \lambda \) scales the effect of \( \sigma^+_v \) to stabilize training.

\section{Experimental Setup}

\subsection{Datasets}
To perform test-time generalization, both synthetic and real datasets have been used.

\begin{table}[t]
\caption{Overview of datasets used.}
\label{tab:t1}
\centering
\footnotesize
\begin{tabular}{llrrr}
\toprule
\textbf{Type} & \textbf{Dataset} & \textbf{\# Images} & \textbf{Image Resolution} & \textbf{\# Classes} \\
\midrule
\multirow{2}{*}{Source} 
    & GTA5 & 24,966 & 1920$\times$1024 & 19 \\
    & Synthia & 9,400 & 1280$\times$720 & 13 \\
\midrule
\multirow{2}{*}{Target} 
    & Cityscapes & 5,000 & 2048$\times$1024 & 19 \\
    & IDD & 1,900 & 1920$\times$1024 & 19 \\
\bottomrule
\end{tabular}
\end{table}

\subsubsection{Synthetic Images (Source Domain)}

Synthetic datasets, GTA5\cite{richter2016playing} and Synthia \cite{ros2016synthia}  are used in the source domain (See \textbf{Table} \ref{tab:t1}). GTA5 contains 24966 images from Grand Theft Auto V and simulates real-world complexities, including weather conditions and times of day. Whereas Synthia is a relatively smaller dataset containing 9400 images. It is curated in controlled environments, but it is less diverse and has lower realism as compared to GTA5. Both the datasets were reclassified into 13 classes.

\subsubsection{Real Images (Target Domain)}

The Cityscapes street view dataset \cite{cordts2016cityscapes} is a collection of 19 annotated classes. The images represent European urban scenes with well-organized road structures and minimal disorder. The IDD \cite{varma2019idd} dataset, on the other hand, is a chaotic and unpredictable representation of the complex environment of India. It captures diverse road scenes with a mix of vehicles, pedestrians, animals and roadside obstacles. The overview of these real street view datasets used in the target domain is presented in \textbf{Table} \ref{tab:t1}.
 For our experiments, we reclassified the 19 classes into 13 classes.
 
\subsection{Training Process}

The model was trained with a batch size of 4 on a single A100 GPU (40GB) to optimize memory use and computational efficiency. The teacher network was optimized using the AdamW algorithm with an initial learning rate of \(1 \times 10^{-5}\), which decayed gradually over 50 epochs, allowing the model ample time to learn robust feature representations from synthetic datasets. Separately, the student network was trained using the AdamW optimizer over 5 epochs with a batch size of 2.

Images were resized to a uniform resolution of 960x720 pixels to ensure consistent input size, with random cropping applied to enhance robustness. Additional data augmentation techniques, including color adjustments and blurring, further improved robustness.

%\subsection{Evaluation Metrics}
%The mean Intersection over Union (mIoU) metric is used to evaluate the performance of SynthGenNet by comparing predicted segmentation masks with actual ground truth annotations. A higher mIoU value indicates better segmentation accuracy, indicating the generalizability of the model from synthetic data to real-world street scene segmentation tasks.

%\subsection{Employed UDA Methods}
%We used four different state-of-the-art domain adaptation methods for our generalization task, including ClassMix\cite{olsson2021classmix}, MIC\cite{hoyer2023mic} , contrastive learning\cite{marsden2022contrastive} and DeepLabv3+\cite{chen2018encoder} into the teacher network and integrated pseudo-label-guided Contrastive Learning\cite{basak2023pseudo} into the student network. As explained in \textbf{Section} ~\ref{sec:SynthGenNet}, we leverage these UDA techniques to generalize from the synthetic source domain to our pseudo label of the target domain. Previous studies have indicated that applying UDA methods to different domain shifts can lead to reduced performance.

\section{Evaluation and Discussion}
\begin{figure*}[!t]
    \centering
    \includegraphics[height=5.5cm, width=11cm]{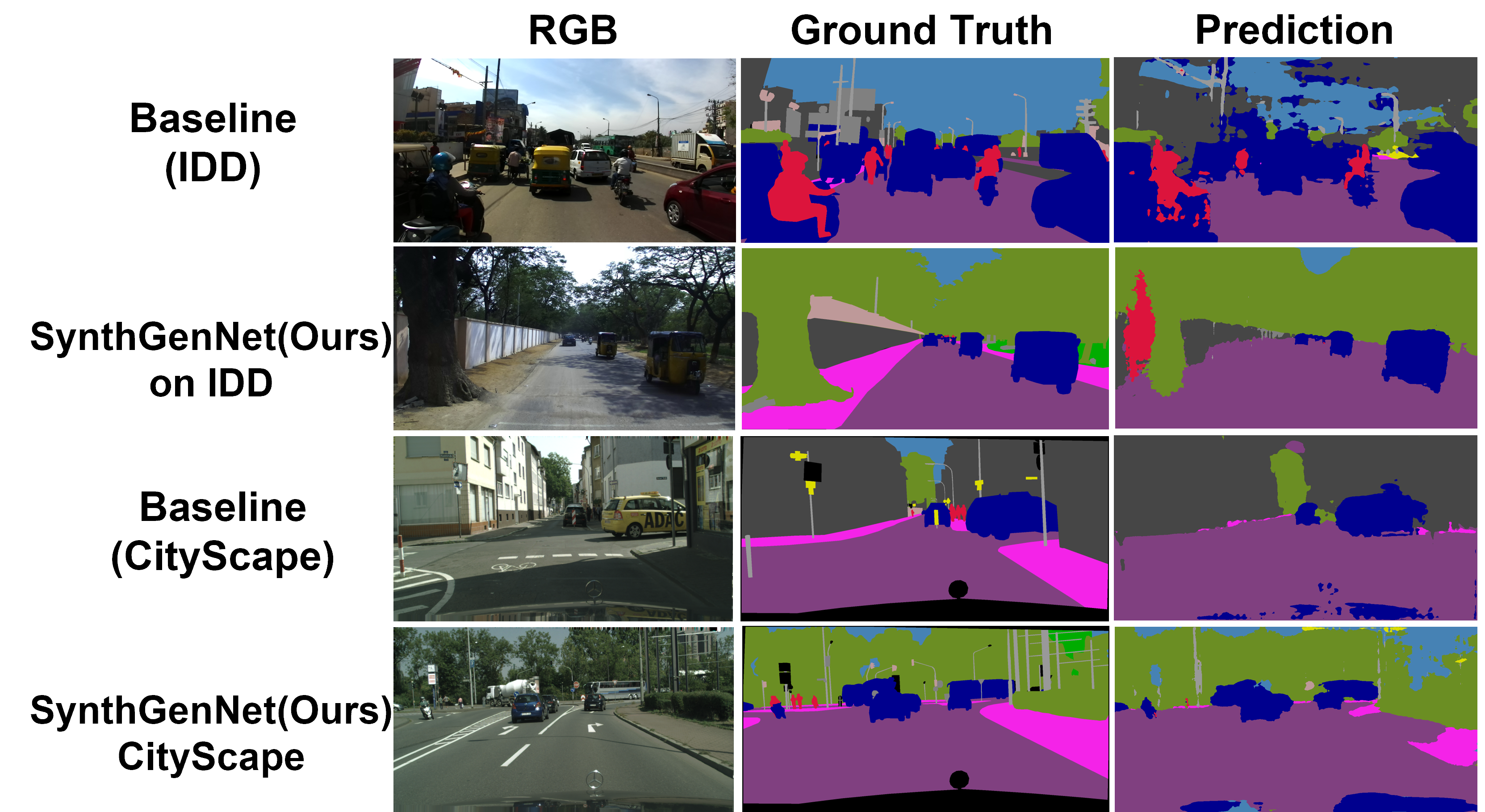}
    \vspace{0.1cm}  
    \caption{Qualitative results of SynthGenNet and Baseline methods on unseen images from IDD and Cityscapes (in row order). SynthGenNet demonstrates superior segmentation accuracy and detail capture in complex urban environments.}
    \vspace{0.1cm} 
    \label{figure 5}
\end{figure*}
This section examines the performance of our proposed architecture \textit{SynthGenNet} and compares it with state-of-the-art methods. For the experiments, there was no pre-existing baseline so to validate the work we have developed a baseline for multi-source domains using the DeepLabV3+ model. This section also explores the impact of techniques like Class-Mix++, GMC loss and PLGCL loss, across different datasets, covering both structured and unstructured environments.
\subsection{Impact of ClassMix++ Integration}
The integration of ClassMix++ improves structured urban scenes like Cityscapes by combining semantic information from different domains, enhancing training data diversity and allowing better generalization. \textbf{Table~\ref{tab:t2}} and \textbf{Table~\ref{table:t0}} show training outcomes on mixed dataset (Smix). The model achieved an excellent mIoU of 41.88\%, outperforming models trained solely on synthetic datasets like GTA5.

This improvement shows the generalizability of ClassMix++ to help maintain semantic consistency and preserve object boundaries, enabling the model to learn spatial relationships for structured and unstructured environments. 

\subsection{Comparison With Existing Methods}
The generalization performance of our method is compared with state-of-the-art approaches in \textbf{Table~\ref{table:t6}}.Our student network model outperforms previous (see \textbf{Table~\ref{table:t3}}), achieving an mIoU of 48.33 on Cityscapes and 49.79 on IDD using the ResNet-50 \cite{he2016deep} backbone whereas 46.63 is achieved with ResNet-101 \cite {he2016deep}. It also surpasses existing methods like TLDR \cite{marsden2022contrastive} and empirical generalization \cite{piva2023empirical}, which were trained and tested on real datasets, particularly on the IDD dataset. \textit{SynthGenNet} shows remarkable performace, for both structured and unstructured environments, especially in handling diverse object types and complex road layouts of developing countries. This generalizability is crucial for real-world applications, such as autonomous driving, where reliable performance must be maintained despite environmental variability. Figure \ref{figure 5} illustrates examples where \textit{SynthGenNet} achieves highly accurate outcomes across a variety of environments.

\subsection{Impact of GMC Loss on Performance}
GMC loss plays a critical role in improving the segmentation of smaller and overlapping objects, such as pedestrians or vehicles partially occluded by other objects. Unlike standard loss functions, GMC forces the network to look beyond local visual cues and incorporate broader contextual information. This leads to more accurate predictions in areas where visual details may be incomplete or unclear (see \textbf{Figure}~\ref{fig-4}). For example, the application of GMC loss, As shown in \textbf{Table~\ref{tab:t2}} and \textbf{Table~\ref{table:t0}}, results in a noticeable improvement in mIoU from 33.56 to 41.88 when applied to Smix. This is particularly beneficial in adverse conditions like lighting or occlusions, where traditional models often fail.  Thus, while GMC enhances robustness.

\subsection{Effectiveness of PLGCL Loss on Performance }
It can be observed that (see \textbf{Table~\ref{table:t4}}) using PLGCL with an optimal weight balance along with cross-entropy loss significantly improves segmentation performance, reflecting high mIoU scores on datasets like Cityscapes and IDD. Using a weight of 1.0 for the contrastive loss and 0.5 for the cross-entropy loss improved generalization across both major and minor classes. This balanced integration helps the network effectively distinguish between these classes, reducing the risk of overfitting to more frequently occurring classes.

The improved performance shows that PLGCL prevents feature collapse and encodes meaningful spatial relationships between patches. It also helps the model to capture fine-grained class differences and learn features that apply across multiple unseen domains. PLGCL is especially valuable in complex scenes such as IDD with diverse object appearances and structures, supporting consistent feature embeddings even in varied environments.

\begin{table}[H]
\centering
\caption{Results of teacher network (ResNet 50)}
\label{tab:t2}
\footnotesize
\begin{tabular}{l l l l}
\toprule
& \textbf{S1} & \textbf{S2} & \textbf{Smix} \\
\midrule
\textbf{Source Training} \\
Seg Loss & 0.09 & - & - \\
MIC Loss & - & 0.17 & 0.17 \\
MIoU     & 0.47 &\textbf{ 0.33} & \textbf{0.41} \\
\midrule
\textbf{Source Validation} \\
Seg Loss & 0.09 & - & - \\
MIC Loss & - & 0.19 & 0.19 \\
MIoU     & 0.47 & 0.34 & 0.41 \\
\bottomrule
\end{tabular}
\end{table}

\begin{table}[H]
\centering
\caption{Results of teacher network (ResNet 101)}
\label{table:t0}
\footnotesize
\begin{tabular}{l l l l}
\toprule
& \textbf{S1} & \textbf{S2} & \textbf{Smix} \\
\midrule
\textbf{Source Training} \\
Seg Loss & 0.10 & - & - \\
MIC Loss & - & 0.18 & 0.17 \\
MIoU     & 0.49 & 0.34 & 0.42 \\
\midrule
\textbf{Source Validation} \\
Seg Loss & 0.09 & - & - \\
MIC Loss & - & 0.09 & 0.11 \\
MIoU     & 0.49 & 0.35 & 0.43 \\
\bottomrule
\end{tabular}
\end{table}

Further validation of \textit{SynthGenNet} highlights its ability to capture fine-grained details. This includes pedestrian outlines and vehicle boundaries. This includes pedestrian outlines and vehicle boundaries. The model performs well even in challenging scenarios like low-light conditions and occlusions. These results show the generalizability of \textit{SynthGenNet} in diverse real-world environments, making it a valuable tool for applications in autonomous driving and urban planning.

\begin{table}[H]
\centering
\caption{Results of student network}
\label{table:t3}
\footnotesize % Adjusting the font size to make the table fit nicely
\begin{tabular}{l l c}
\toprule
\textbf{Model} & \textbf{Dataset} & \textbf{MIoU} \\
\midrule
\multirow{2}{*}{\textbf{ResNet 50}} & Cityscape & 48.33 \\
                                     & IDD       & \textbf{49.79 }\\
\midrule
\multirow{2}{*}{\textbf{ResNet 101}} & Cityscape & 48.87 \\
                                     & IDD       & \textbf{46.63 }\\
\bottomrule
\end{tabular}
\end{table}

\section{Conclusion}
We present \textit{SynthGenNet}, a novel test-time generalization architecture optimized for the complexities of unstructured urban environments, which utilizes multi-source synthetic data, ClassMix++, and GMC loss to deliver superior segmentation across diverse domains. Surpassing existing benchmarks with a 49.79 mIoU on the IDD dataset, our architecture demonstrates improved generalization capabilities, capturing intricate segmentation details even under challenging real-world conditions. Although the computational demands of GMC loss present constraints for real-time deployment, future work can focus on optimizing this component to enhance efficiency. Additionally, exploring transformer-based models with adaptive contrastive loss offers a promising path for further performance gains. 
\begin{table}[H]
\centering
\caption{Effectiveness of contrastive loss on performance}
\label{table:t4}
\footnotesize % Adjusted text size
\begin{tabular}{cccc}
\toprule
\textbf{Encoder} & \textbf{Embedding Size} & \textbf{\begin{tabular}[c]{@{}c@{}}Cross Entropy \\ Loss\end{tabular}} & \textbf{MIoU} \\
\midrule
\multirow{3}{*}{ResNet 50} & \multirow{3}{*}{1024} & 0.5 & 41.51 \\
 & & 1.0 & \textbf{49.79} \\
 & & 0.5 & 31.57 \\
\bottomrule
\end{tabular}
\end{table}

\begin{table}[H]
\centering
\caption{Test-time domain generalization performance (mIoU, \%) of various methods using ResNet-50 and ResNet-101 encoder networks. Training is conducted on multiple synthetic source datasets (GTA5 and SYNTHIA), with evaluation performed on real-world validation sets (IDD and Cityscape).}
\label{table:t6}
\footnotesize % Use a smaller text size
\begin{tabular}{clcc}
\toprule
\textbf{Model} & \textbf{\begin{tabular}[c]{@{}c@{}}Training\\ Domain\end{tabular}} & \textbf{Cityscape} & \textbf{IDD} \\
\midrule
\multicolumn{4}{c}{\textbf{Encoder: ResNet 50}} \\
Baseline & GTA5, Synthia & 30.57 & - \\
TLDR \cite{marsden2022contrastive} & GTA5 & 46.51 & - \\
Empirical Generalization \cite{piva2023empirical} & Cityscape & - & 48.7 \\
\textit{SynthGenNet }(Ours) & GTA5, Synthia & \textbf{48.33} & \textbf{49.79 }\\
\midrule
\multicolumn{4}{c}{\textbf{Encoder: ResNet 101}} \\
Baseline & GTA5, Synthia & 31.87 & - \\
TLDR \cite{marsden2022contrastive}  & GTA5 & 47.58 & - \\
\textit{SynthGenNet} (Ours) & GTA5, Synthia & \textbf{48.87} &\textbf{ 46.33} \\
\bottomrule
\end{tabular}
\end{table}

% ====== REFERENCES ======
\bibliographystyle{unsrt} % or plain, ieeetr, alpha, etc.
\bibliography{main}  % your BibTeX file

\end{document}